\pdfoutput=1

\documentclass[11pt]{article}

\usepackage[final]{acl}

\usepackage{times}
\usepackage{latexsym}

\usepackage[T1]{fontenc}

\usepackage[utf8]{inputenc}

\usepackage{microtype}

\usepackage{inconsolata}

\usepackage{graphicx}

\usepackage{booktabs}       
\usepackage{amsfonts}       
\usepackage{tabularx} 
\usepackage{caption} 
\usepackage{multicol}
\usepackage{multirow}
\usepackage{array}
\usepackage{amsmath}
\usepackage{comment}
\usepackage{algorithm}
\usepackage{algpseudocode}

\newcommand{\eat}[1]{}

\title{Improve Dense Passage Retrieval with Entailment Tuning}

\author{
  \textbf{Lu Dai}\textsuperscript{1},
  \textbf{Hao Liu}\textsuperscript{1,2},
  \textbf{Hui Xiong}\textsuperscript{1,2}
\\
  \textsuperscript{1}Thrust of Artificial Intelligence, \\
  The Hong Kong University of Science and Technology (Guangzhou), China \\
  \textsuperscript{2}Department of Computer Science and Engineering, \\
  The Hong Kong University of Science and Technology \\
  Hong Kong SAR, China
\\
{\tt\small ldaiae@connect.ust.hk} {\tt\small \{liuh,xionghui\}@ust.hk}
}

\begin{document}
\maketitle

\begin{abstract}
Retrieval module can be plugged into many downstream NLP tasks to improve their performance, such as open-domain question answering and retrieval-augmented generation. 
The key to a retrieval system is to calculate relevance scores to query and passage pairs. However, the definition of relevance is often ambiguous. We observed that a major class of relevance aligns with the concept of entailment in NLI tasks. Based on this observation, we designed a method called entailment tuning to improve the embedding of dense retrievers. Specifically, we unify the form of retrieval data and NLI data using existence claim as a bridge. Then, we train retrievers to predict the claims entailed in a passage with a variant task of masked prediction. 
Our method can be efficiently plugged into current dense retrieval methods, and experiments show the effectiveness of our method. 
\end{abstract}
\section{Introduction}

Information Retrieval(IR) is the process of searching and matching relevant information for a given query. Due to its effectiveness, IR has been integrated into a wide range of modern NLP solutions, especially knowledge-intensive tasks such as open-domain QA~\cite{dpr2020, drqa2017}, fact verification\cite{fever2018} and retrieval-augmented generation (RAG)~\cite{realm2020, rag2020}. With the development of pre-trained language models (PLM), dense retrieval methods
~\cite{condenser2021, retromae2022} have demonstrated remarkable performance in IR tasks by matching queries and contexts using vector representations learned by PLMs. In such a way, texts can be retrieved based on semantic relevance, thus avoiding problems such as vocabulary mismatching and providing more useful information for downstream tasks~\cite{ance2020}.

A key challenge to IR is the ambiguous definition of relevance~\cite{survey:rag2024}. On the one hand, the exact meaning of relevance varies across different tasks and user intents~\cite{instructor2023}. For example, while defining relevance as keyword matching adequately addresses most needs of general web searches, it is insufficient for open-domain question answering, where the relevance of a text hinges on whether answers can be logically deduced from it. On the other hand, while dense retrievers are built upon PLMs, there exists a nuanced gap in how they define relevance, owing to their differing training objectives~\cite{bridgegap2024}. PLMs like BERT~\cite{bert2019} are trained under masked token prediction. As a result, texts co-occurred in the same context window has similar representations, which misaligns with the target of retrieval systems. 

\begin{figure}[t]
    \centering
    \includegraphics[width=\linewidth]{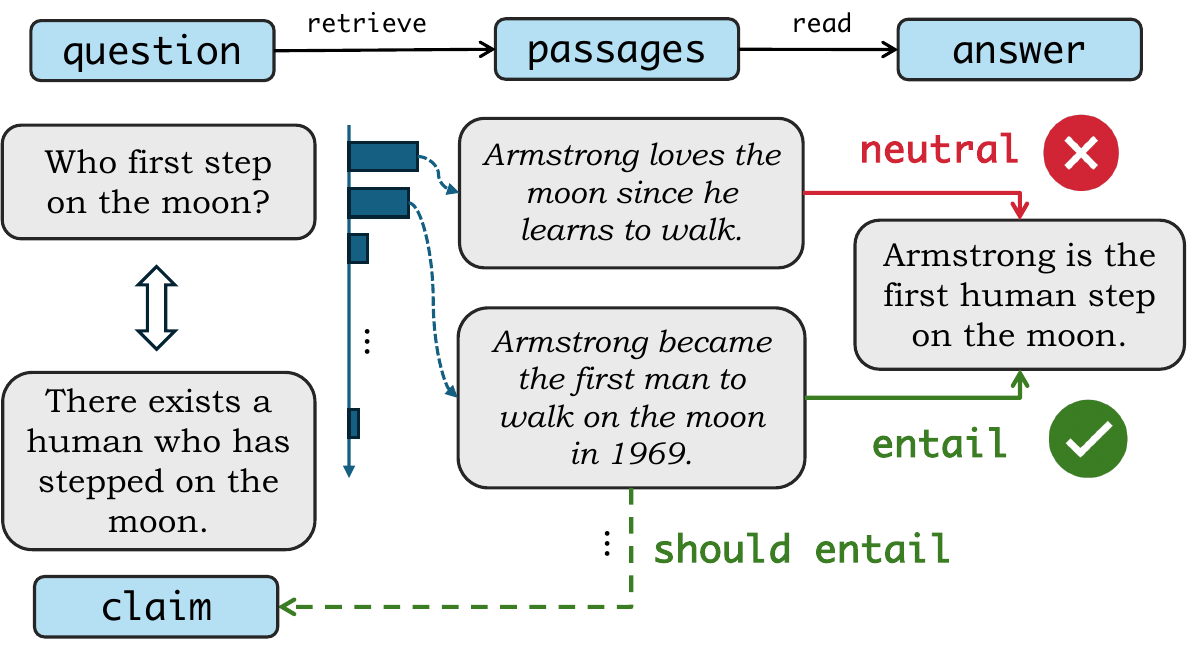}
    \caption{Both passages contains answer and receive high relevance score, but only the second is truly helpful to deduce answer. A necessary condition of a helpful passage is entailing the claim underlying the question.}
    \label{fig:enter-label}
\end{figure}

Several works has managed to improve retrieval by covering different aspects of relevance~\cite{dorismae2024, polyencoder2019} and tailor retrieval schemes for different tasks~\cite{scincl2022, lawnli2022, lawrelevance2017, instructor2023}. However, they are largely data-driven or confined to specific domain and lacks a clear and versatile definition of the relevance for effective retrieval. More advanced understanding of relevance in retrieval can help to break the quality bottleneck of downstream tasks such as RAG.

In this work, we reconsider the relevance definition in QA-oriented retrieval from the lens of natural language inference(NLI)~\cite{stsfromnli2017}. For example, consider the query "Who first step on the moon?", the actual information flow into a retrieval system is "there exists a human being who has already stepped on the moon sometime." Thus, a necessary condition for a positive passage which is relevant and helpful to answer this question is that the claim can be logically inferred from it. Using the framework of NLI, we observed that when regarding passages as premise and claim as hypothesis, the relationship between positive/relevant passages and claim is entailment, while the relationship between negative/irrelevant passages and claim is neutral. If the claim itself is not held true, then retrieved passages might even contradict the claim, in which case the question is defined as unanswerable. To validate this formulation, we conduct several experiments on the correlation of relevance and entailment. We show that off-the-shelf NLI models indeed tend to assign significantly higher entailment probability to positive passages, and dense retrievers also give higher relevance scores to the premise and hypothesis of the entailment relationship compared to neutral and irrelevant ones. 

Based on the above formulation and empirical evidence, we designed a method called entailment tuning to enhance the performance of dense retrievers for open-domain QA. We augment dense retriever training with NLI data~\cite{dataset:snli2015, dataset:mnli2018}, and draw closer the embedding of text pairs of entailment relationship. Specifically, we first convert questions to claims using a rule-based method and assemble claim-passage pairs and premise-hypothesis pairs in a unified prompt. Then, we mask almost the whole span of hypothesis part and train the encoder model to predict the masked hypothesis from the premises. This encourages the passage embedding to focus on the information it entails, and consequently improve retrieval performance at inference time. Our experiments demonstrate the effectiveness of the entailment tuning method in dense passage retrieval tasks.

In summary, our contribution is three-fold: first, we propose and validate a perspective of defining query-passage relevance using the concept of entailment from NLI. Second, by exploiting this connection, we design an algorithm called entailment tuning which can be easily plugged into SOTA dense retriever training pipelines, and empirically validated its significant effectiveness in vast amount of datasets and methods. Third, We further verify that enhancing the entailment type of relevancy in retrieved passages indeed translates to better performance in the downstream tasks of retrieval, such as open-domain QA and RAG.
\section{Background}

\begin{table*}[ht]
\centering
\captionsetup{skip=10pt}  
\begin{tabular}{p{9cm}cc}
\toprule
\textbf{Example (Passage \& Query)} & \textbf{Relationship} & \textbf{Downstream Task} \\
\midrule
\begin{tabular}{@{}p{9cm}@{}}
\scriptsize\textbf{Passage:} The Great Lakes, also called the Laurentian Great Lakes and the Great Lakes of North America, are a series of interconnected freshwater lakes located primarily in the upper mid-east region of North America, on the Canada–United States border, which connect to the Atlantic Ocean through the Saint Lawrence River. \\
\scriptsize\textbf{Query:} Where do the Great Lakes meet the ocean?
\end{tabular} & \small entail & \small Open-domain QA \\
\midrule
\begin{tabular}{@{}p{9cm}@{}}
\scriptsize\textbf{Passage:} While I do agree that there are emotionally and physically demanding aspects to both dance and sports, there are too many differences between them to call dance a sport itself. For example, Dance exists to tell a story through movement and music. That is something sports simply do not do.  \\
\scriptsize\textbf{Query:} Dance Is Not a Sport
\end{tabular} &  \small entail, contradict & \small Argument Retrieval \\
\midrule
\begin{tabular}{@{}p{9cm}@{}}
\scriptsize\textbf{Passage:} Elon Musk hires a team of experts to build the ultimate yacht, but when the yacht is completed, he realizes that he has no idea how to sail it. With the help of a quirky crew and a fearless captain, the playboy embarks on a wild and hilarious adventure across the open seas, where the crew have to keep Elon alive despite his inability to do anything himself. All the while, Elon takes credit for their hard work. \\
\scriptsize\textbf{Query:} Write a plot summary for a comedic novel involving Elon Musk and sea travel.
\end{tabular} & \small constraint satisfaction & 
\begin{tabular}{@{}c@{}}
\small RAG \\ 
\small (instruction-following)
\end{tabular}\\
\bottomrule
\end{tabular}
\caption{Examples of different meaning or relevance between passage and query in retrieval-related tasks. While QA seeks passages that entails the information in query, argument retrieval tasks seek passages both entails and contradict the query. RAG covers a wider range of relevance definition, such as constraint satisfaction.}
\label{tab:relevance_type}
\end{table*}

\textbf{Dense Retrieval}
Unlike traditional methods such as TF-IDF and BM25\cite{bm252009} that calculate text relevance based on term frequency, dense retrieval methods use deep neural networks to encode a piece of text which integrates contextualized information into a single vector, and then text is retrieved with maximum inner-product search (MIPS) based on its embedding similarity with query. Siamese network and dual encoder are the two most frequently used structures to encode queries and passages, which are built on the PLMs with advanced language understanding abilities. However, general-purpose PLMs are not trained under retrieval objectives. To this end, lines of methods have been proposed to adapt LMs to retrieval tasks. At pre-training stage, several works\cite{retromae2022, condenser2021, contriever2022} design unsupervised training schemes suitable for retrieval task, including aggressive masking, asymmetric encoder-decoder, and inverse-cloze task\cite{ict2019}. At fine-tuning stage, techniques like supervised contrastive learning\cite{dpr2020, ance2020, rocketqa2021} and late-interactions\cite{colbert2020} are used to train dense retrievers. With recent developments of GPT-like models, another line of work attempts to adapt LLMs for IR tasks, using methods like bi-directional attention\cite{llm2vec2024}, instruction-tuning\cite{tart2023, instructor2023} and hypothesis generation\cite{hyde2023}. \\
\textbf{Natural Language Inference}
Natural language inference is a fundamental task in NLP, underpinning a wide range of NLP tasks, from commonsense reasoning to semantic textual similarity (STS) tasks\cite{dataset:snli2015, stsfromnli2017}. NLI focuses on understanding sentence meaning and the relationship between sentences. Specifically, given a premise sentence and a hypothesis sentence, the goal is to classify the relationship of the two sentences into three categories: \textit{entail}, \textit{neutral} or \textit{contradict}. \textit{Entail} means the hypothesis can be logically inferred from information provided in the premise. \textit{Neutral} means hypothesis can not be deduced conditioned on premise, although they may have a large topic or lexical overlap. \textit{Contradict} means the if the premise stands true, then the hypothesis must be false. Challenge in NLI task lies in an accurate understanding of deep semantic meaning beyond shallow features of natural languages\cite{dataset:mnli2018, bertflow2020, use2018}. Recently, researchers found that using NLI data for supervised training benefits learning sentence embeddings thus improving the performance of downstream tasks such as sentimental analysis and opinion polarity detection. Earlier work such InferSent\cite{stsfromnli2017} learn sentence embedding based on LSTMs or CNNs\cite{cnnsts2014}. Utilizing the representation power of PLMs, models like SBERT and SimCSE\cite{sbert2019, st52022, simcse2021} further proves the feasibility of improving STS tasks using NLI data as supervision signal. However, these exploration mainly works the grain of sentences and a feasible end-to-end solution to be applied to dense passage retrieval remains lack.


\section{Preliminaries}
In this section, we introduce the preliminaries of current dense retrieval framework.

\noindent\textbf{Task Definition.} Given a query $q$ and a collection of passages $\mathcal{P}$, the goal of the dense retriever $M$ is to retrieve $k$ passages from $\mathcal{P}$ that are most relevant to $q$. Passages in $\mathcal{P}$ are encoded using $M$ and represented in the form of dense vectors. They are pre-calculated and stored in a vector database, organized using a index such as FAISS\cite{faiss2019}. At inference time, query $q$ is first encoded using $M$. Then, the relevance score of query $q$ and a passage $p$ is calculated using a similarity function $f$, and $k$ passages with highest scores are retrieved and returned:
\begin{equation}
\{ p_1,\ldots,p_k \} =\underset{k}{\text{argsort}}\ f(M(q;\theta), M(p;\theta))
\end{equation}
Training of denser retriever $M$ can be broadly classified into two stages, which are instruced as follows.

\noindent\textbf{Pre-training.} Based on PLMs, dense retriever are further trained in an unsupervised manner on large-scale corpus such as NQ and MSMARCO. This retrieval-oriented pre-train aims to adapt PLMs to dense retrieval by encoding richer information into document embedding, using tasks such as inverse cloze prediction and Maskedsalient spans. This stage is not necessary but can provide better initialzation for fine-tuning.

\noindent\textbf{Fine-tuning.} Fine-tuning employs supervised training scheme using much smaller annotated retrieval data. The paired dataset $\mathcal{D}$ consists of triplets $\{ (q_i, p_i^+, p_{i,1}^-, \ldots, p_{i,m}^-) \}_{i=1}^n$, where $q$ is the query, $p^+$ is a positive passage relevant to $q$ and $\{p_1^-,\ldots, p_m^-\}$ are negative passages irrelevant to $q$. Dense retriever is often trained with a constrastive loss funtion as follows:
\begin{equation}
    L_{nll} = - \mathbb{E}_\mathcal{D} \left[ \log \frac{e^{\text{sim}(q_i, p_i^+)}}{e^{\text{sim}(q_i, p_i^+)} + \sum_{j=1}^n e^{\text{sim}(q_i, p_{i,j}^-)}} \right]
\end{equation}

Our entailment tuning method can be easily plugged into current dense retrieval pipeline between pre-train and contrastive fine-tune. It is simple and efficient, which costs only a overhead on the two main stages. Details of entailment tuning is described in Section~\ref{sec:entail}.
\section{Rethinking relevance in retrieval-augmented QA}
\label{sec:3}
\subsection{Different types of relevance}
Relevance scoring is crucial for selecting and ranking passages in retrieval tasks. The definition of relevance, however, varies depending on the specific requirements of the task and user intent. For example, relevance in news retrieval is often based on topical or lexical similarity, ensuring that content matches the search theme. Open-domain question-answering (QA) tasks, on the other hand, demand that relevance be tightly defined as containing precise information necessary to answer the query. In argument retrieval tasks, relevance includes both supporting and opposing passages.

In Table ~\ref{tab:relevance_type}, we provide examples to highlight the different interpretations of relevance across these contexts. Although defining relevance in complex tasks remains challenging, we find a perspective based on NLI to model the relationships between passage and answer and demonstrate its effect.

\subsection{Question and Existence Claim}

When an inquirer poses a question, they typically possess some foundational knowledge about the subject but face uncertainties regarding specific details. For example, consider the question: \textit{When was the movie Titanic released?} 
Information contained in this question is that the event "release of the Titanic movie" occurred at a determinable time in the past.

\noindent\textbf{Definition 1:} \textit{An \textbf{existence claim} $c$ is a logical statement in the pattern of:}
\begin{equation}
\exists x : \mathcal{C}(x),
\end{equation}
where \( \mathcal{C}(x) \) is a predicate expressing the occurrence or presence of an event or entity \( x \).
Similarly, we can reformulate any question into an existence claim $c$ without loss of information, if the question itself is valid:\\
\noindent\textbf{Proposition 1:} \textit{A valid $q(x)$ can be transformed to $c$ in an information-invariant way:}
\begin{equation}
q(x)\ \text{is vaild} \Leftrightarrow \exists x : \mathcal{C}(x)
\end{equation}
In retrieval, desired relevant passage $p$ should be texts from where answers can be logically inferred:
$p \rightarrow a(q)$, where $p$ represents a passage, and $a$ represents the answer to $q$. This logical form, known as \textit{entailment}, is crucial in logic and natural language inference (NLI) tasks.
 
\noindent\textbf{Proposition 2 (Chain Rule):} \textit{The logical interdependence of statements can be encapsulated in the chain rule, which states:}
\begin{equation}
(p \rightarrow a) \wedge (a \rightarrow c) \rightarrow (p \rightarrow c).
\end{equation}
This theorem illustrates a fundamental principle in logic: if passage $p$ entails answer $a(q)$, and $a(q)$ in turn entails existence claim $c$, then it follows that $p$ indirectly entails $c$. Thus, that $p$ entails $c$ is a neccessary (but not sufficient) condition of that $p$ entails $a(q)$, i.e.\, $p$ is relevant to $q$.

Given that during the inference stage of a retrieval system, the answer $a(q)$ is unknown, we use existence claim $c$ as a lower-bound of $a(q)$ and minimize the distance between a passage and its corresponding existence claim in the embedding space. By optimizing this relationship, we enhance the scoring of passages which can truly deduce the answer to a question. 

\begin{figure}[h]
  \centering
  \includegraphics[width=0.48\textwidth]{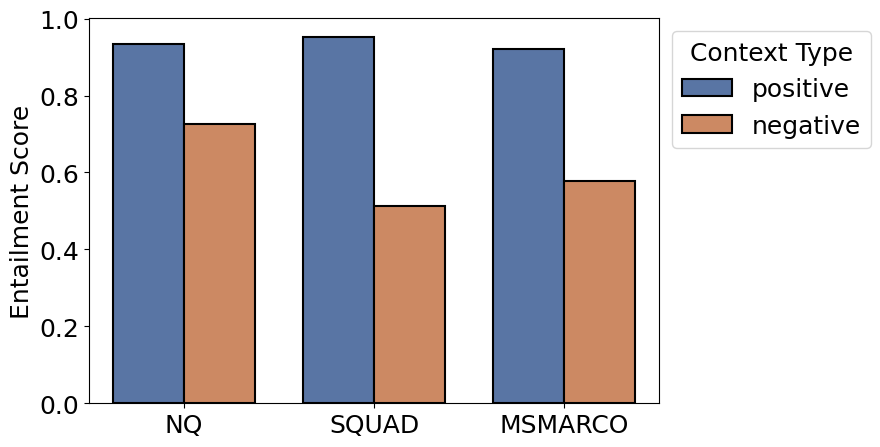}
  \caption{NLI model has a clear tendency to classify the relationship between possitive passage and query as entailment, compared to negative passages and query.}
  \label{fig:nli_score}
\end{figure}

\subsection{Retrievers and NLI models}

One immediate question is whether NLI models can discern the relationship wherein a passage entails a claim during retrieval tasks. We conducted experiments using a robust NLI model based on RoBERTa, testing it across three distinct datasets: NQ, SQuAD, and MS MARCO. We classified passages as "positive" if the answers could be inferred from them, and as "negative" if they did not contribute to finding the answer. We input the passages and claims into the NLI models as premises and hypotheses, respectively, and received a score indicating the probability that the premise entails the hypothesis. The results, presented in Figure ~\ref{fig:nli_score}, show that the NLI model consistently assigns higher probabilities to positive passages, suggesting that they entail the claims. This distinction is pronounced when compared to the scores assigned to negative passages.

\begin{figure}[h]
    \centering
    \includegraphics[width=\linewidth]{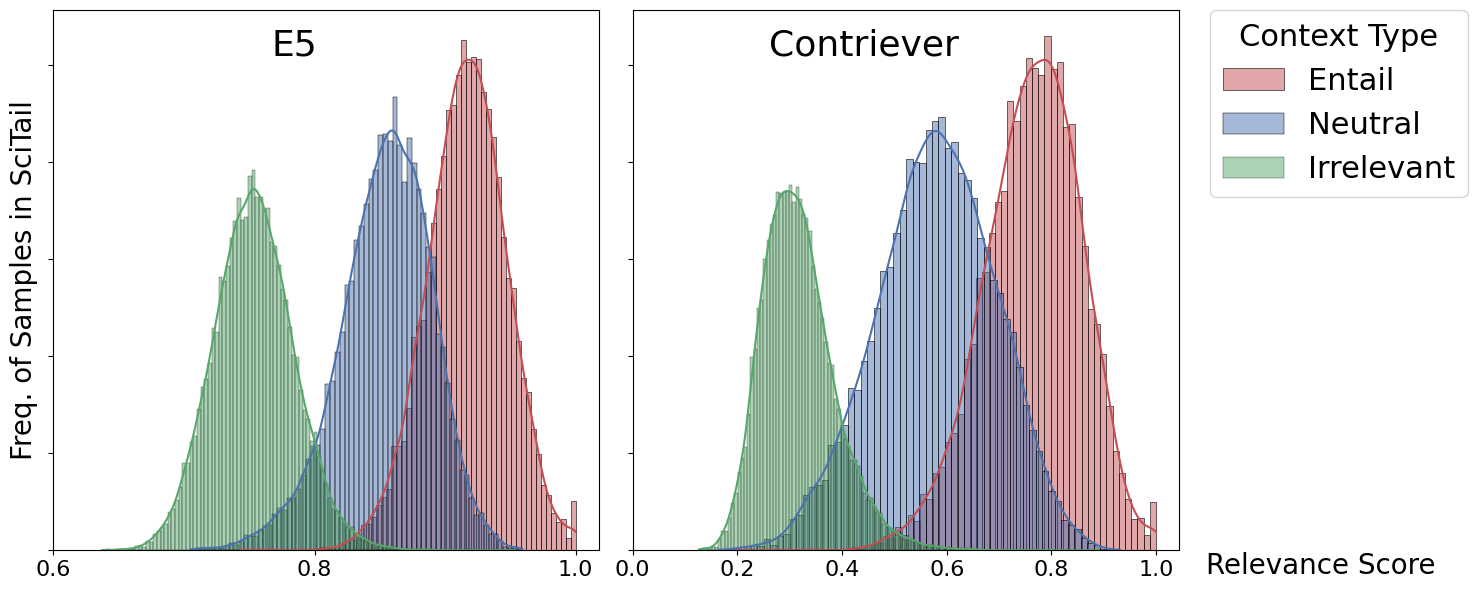}
    \caption{Dense retriever can discern sentence pairs of different semantic relationships, shown by separate relevance score range, especially entail and irrelevant, but still has some difficulty between entail and neutral.}
    \label{fig:retriever_score}
\end{figure}

Furthermore, we examined the capability of retrieval models to differentiate between passages that entail an answer and those that do not. For each hypothesis, we selected three types of premises: entail, neutral, and irrelevant. An 'entail' premise directly supports the hypothesis, a 'neutral' premise shares significant topical or lexical overlap without supporting the hypothesis, and an 'irrelevant' premise consists of discourse randomly sampled from the corpus, aligning with the general definition of irrelevance. The findings, detailed in Figure ~\ref{fig:retriever_score}, indicate that current retrieval models effectively distinguish irrelevant from entail content. However, they struggle to differentiate between entail and neutral premises. This issue is particularly problematic in challenging retrieval scenarios, where the content, despite its topical relevance, fails to provide actionable insights for answering the query. Such scenarios pose significant obstacles for downstream tasks like RAG and QA, representing a persistent challenge for contemporary retrieval systems.

\section{Entailment Tuning}
\label{sec:entail}
In this section, we introduce our method of tuning dense retrievers by enhancing the entailment relationship between query and retrieved passages. This method can be easily plugged into current dense retriever training pipeline before supervised contrastive finetuning. \\
\label{unified_prompt}
\textbf{Unified Prompting.} To use NLI data to augment the entailment tuning process, we first unified the format of NLI data and passage retrieval data. 

NLI data consists of pairs of statements premise and hypothesis. We use the prompt \textit{"<premise> entails that <hypothesis>"} to assemble the pair in our entailment tuning process. Passage retrieval data, on the other hand, consists of triples of $(q, p^+, p^-)$, where $p^+$ stands for positive passages, and $p^-$ stands for negative passages. To fit passage retrieval data into the entailment prompt, the question $q$ is first transformed into a narrative-form claim $c$ in an information-variant manner. 

Specifically, we use a set of rules to effectively convert $q$ to $c$. We divide questions into six categories: \textit{When, Why, Who, Where, Does, How}. For example, a question in the form of "when did ..." is then mapped into claim "There exists a known time when ...". Then, $p^+$ is positioned in \textit{<premise>} and $c$ is put into \textit{<hypothesis>} which can be deduced from $p^+$. In this way, the passage data $(q, p^+)$ can be composited in the same format like other NLI data and mixed together in training. \\

\label{masked_hytpothesis}
\noindent\textbf{Masked Hypothesis Prediction.} Once we get the unified formatted data, we adapted the masked-prediction scheme in our entailment tuning setting.

Like all general MLMs, we masked part of the prompt sentence and requires the model to predict the masked tokens. 
Unlike other MLMs that randomly choose tokens to mask, we mask almost the whole \textit{<hypothesis>} part and leave the \textit{<premise>} part visible. 

Given a premise $P$ and a hypothesis $H = h_1, h_2, \dots, h_n$, each token $h_i$ in $H$ is independently masked with a probability $\beta$ which is much higher than MLM pre-training. The masked hypothesis $H_{\text{masked}}$ is formed by replacing each token $h_i$ with a token $m_i$, where:
\[
m_i = 
\begin{cases} 
[\text{MASK}] & \text{with probability } \beta \\
h_i & \text{with probability } 1 - \beta
\end{cases}
\]
The input to the model is then defined as:
\begin{equation}
X = Prompt(P,H_{\text{masked}})
\end{equation}
Then, $M$ is trained using a masked prediction objective:
\begin{equation}
    \mathcal{L}_{mlm} = -\sum_{i \in\{\text{[MASK]\}}} \log P(\hat{h}_i = h_i | X)
\end{equation}

We design this scheme based on several intuition and evidences. First, since premise entails hypothesis, the premise should contain sufficient information to predict hypothesis. Second, long-range masking improves the global representation ability of language model\cite{t52020, retromae2022, shouldmask2023}. In BERT, around 15\% tokens are randomly masked. In this way, there are always large portion of unmasked tokens around one single masked token, which encourages the model to learn word or phrase level local embedding. On the contrary, our model mask a continuous long span in the sentence, which impels the model to aggregate global information in premise to correctly predict premise. Third, we specifically mask the hypothesis part. This encourages the model to engrave the information entailed in the premise into its embedding. In this way, a model can retrieve passages that has an entailment relationship with input query, which is of higher quality according to our analysis in Section~\ref{sec:3}.

\begin{algorithm}
\caption{Entailment Tuning}
\begin{algorithmic}[1]
    \State \textbf{begin}
    \State $W_M \gets W_{BERT}$  \Comment{\textit{Initialize model}}
    \If{$\text{data} \in D_{\text{retrieval}}$}\Comment{\textit{Transform data}}
        \State $P \gets p^+$, $H \gets c$
    \ElsIf{$\text{data} \in D_{\text{NLI}}$}
        \State $P \gets \text{Premise}$, $H \gets \text{Hypothesis}$
    \EndIf
    \State $H_\text{{masked}} \gets Mask(\beta)$ \Comment{\textit{Mask hypothesis}}
    \State $X \gets Prompt(P, H_\text{{masked}})$ \Comment{\textit{Ensemble input}}
    \For{$\text{epoch} = 1 \ to \ n$} \Comment{\textit{Train}}
        \State $\tilde{H} = M(X)$
        \State $L_{\text{mlm}} = -\sum \log P(\tilde{H})$ 
        \State $W_M = W_M - \eta \nabla L_{\text{mlm}}$
    \EndFor
    
    \State \textbf{end}
\end{algorithmic}
\end{algorithm}

\section{Experiments}
\begin{table*}[ht]
    \centering
    \captionsetup{skip=10pt}  
    \begin{tabularx}{\textwidth}{l|*{5}{X}|*{2}{X}}
        \toprule
        Model & \multicolumn{5}{c|}{\textbf{NQ}} & \multicolumn{2}{c}{\textbf{MSMARCO}} \\
        \midrule
         & R@1 & R@5 & R@20 & R@100 & MRR & MRR@10 & Recall@1K \\
        \midrule
        BM25 & 23.9 & 45.9 & 63.8 & 78.9 & -- & 24.0 & 81.4 \\
        \midrule
        BERT & 45.21 & 68.20 & 79.61 & 86.07 & 64.51 & 31.26 & 95.23 \\
        + \textit{Ent. T.} & \textbf{48.53} & \textbf{70.08} & \textbf{80.94} & \textbf{86.43} & \textbf{67.24} & \textbf{31.89} & \textbf{95.87}\\
        \midrule
        RoBERTa & 43.07 & 66.40 & 77.45 & 84.88 & 62.75 & 29.17 & 94.57 \\
        + \textit{Ent. T.} & \textbf{45.24} & \textbf{66.76} & \textbf{78.56} & \textbf{85.68} & \textbf{64.24} & \textbf{29.97} & 95.02 \\
        \midrule
        RetroMAE & 47.95 & 70.89 & 82.11 & 87.80 & 66.12 & 34.54 & 97.51\\
        + \textit{Ent. T.} & \textbf{49.53} & \textbf{72.02}& \textbf{82.27} & \textbf{87.80} & \textbf{67.75} & \textbf{34.61} & \textbf{97.54}\\
        \midrule
        Condenser & 47.62 & 70.53 & 80.64 & 87.01 & 66.34 & 32.64 & 96.49 \\
        + \textit{Ent. T.} & \textbf{49.75} & \textbf{71.47} & \textbf{81.52} & \textbf{87.29} & \textbf{67.89} & \textbf{33.39} & \textbf{96.62}\\
        \bottomrule
    \end{tabularx}
    \caption{Performance comparison of different models on the NQ and MSMARCO w/ and w/o entailment tuning. Ent. T. means our entailment tuning method is applied to the training pipeline of corresponding dense retriever.}
    \label{tab:retrieval}
\end{table*}

In this section, we evaluated the performance of entailment tuning in passage retrieval, as well as two downstream tasks of open-domain QA and retrieval-augmented generation. We also test its compatibility with different architectures, pretrain schemes and model sizes in previous dense retrieval works. We can see that in tasks where query and context has a relationship that can be captured by entailment, both retrieval and downstream performance consistently outperforms methods that are not equipped with our entailment tuning method. 

\subsection{Passage retrieval}
We use Wikipedia corpus as the pool for retrieval, and test passage retrieval performance using Natural Question (NQ) dataset\cite{dataset:nq2019}.
the most widely used dataset in open-domain QA. 
We implement dense retrieval on the corpus of Wikipedia and MSMARCO, using Natural Question (NQ) dataset\cite{dataset:nq2019} and MSMARCO Dev respectively as test dataset. These are the two most widely used corpus and test setting in dense retrieval.

We insert entailment-tuning between current pre-training/fine-tuning stages for dense retriever training. For the entailment-tuning stage, we use a dataset combination of NQ/MSMARCO, SNLI and MNLI for entailment tuning. We tune PLMs for 10 epochs with a learning rate 2e-5 and batch size 128 on 8 GPUs. For the contrastive fine-tuning stage which is not our contribution, we follow the exact same hyperparameter setting as DPR, elaborated in Appendix. Methods are evaluated with top-k hits and mean reciprocal rank (MRR) metrics in NQ. MRR is abbreviated of MRR@100 following previous works. For MSMARCO, we use MRR@10 and Recall@1K to align with previous works.
To compare the methods compatibility with different models, we choose most widely used and well-performed dense retrievers, BERT\cite{bert2019}, RoBERTa\cite{roberta2019}, DeBERTa\cite{deberta2020}, Condenser\cite{condenser2021} and RetroMAE\cite{retromae2022}. The latter two are specially trained with large scale retrieval-oriented unsupervised pre-training.

We show in Table~\ref{tab:retrieval} that dense retrievers that employed entailment tuning consistently outperforms corresponding baselines, and achieves 1\% to 3\% improvement in top-k hits and MRR. We also noticed two tendency based on experiment results. First, our method brings in higher performance increase in smaller K. For example, compared to DPR, our method improves top-1 hits by 3.32\%, but only improves top-100 hits by 0.36\%. This results suggested that with entailment tuning, the model might become more confident with positive passages where answers can really be deduced from. Second, our method brings in higher improvement for PLMs without retrieval-oriented pre-training. For example, it improves the MRR of dense passage retriever which is based on original BERT by 2.73\%, but improves the MRR of Condenser and RetroMAE by around 1.6\%.
This observation suggests that entailment tuning shares parts of common objectives with these retrieval-oriented pre-training techniques. However, the entailment tuning method is far more efficient by leveraging the power of paired NLI data, compared to pre-training methods which is based on unsupervised training on large-scale data. The entailment tuning process costs less than 2 hours on 8 GPUs, while retrieval-oriented pre-training generally costs around 3 days.

\subsection{Open-Domain QA}

\begin{table}[h]
  \centering
  \begin{tabular}{cc|cc}
    \toprule
    Reader & Method & NQ & TriviaQA  \\
    \midrule
    \multirow{2}{*}{T5$_{\text{base}}$} & w/o Ent. T. & 44.93 & 46.42  \\
                               &  w/ Ent. T. & \textbf{45.43} & \textbf{46.85}  \\
    \midrule
    \multirow{2}{*}{T5$_{\text{large}}$} & w/o Ent. T. & 50.28 & 52.97 \\
                                &  w/ Ent. T. & \textbf{51.36} & \textbf{53.08}  \\
    \bottomrule
  \end{tabular}
  \caption{EM for QA on NQ and TriviaQA datasets.}
  \label{tab:odqa}
\end{table}

\begin{table*}[ht]
    \small
    \centering
    \begin{tabular}{cccccccc}
        \toprule
        \multirow{2}{*}{Generator} & \multirow{2}{*}{Ent. T.} & \multicolumn{3}{c}{\textbf{ELI5}} & \multicolumn{3}{c}{\textbf{ASQA}} \\ \cmidrule(lr){3-5} \cmidrule(lr){6-8}
                                  &  & ROUGE-L & Correctness & Relevancy & ROUGE-L & Correctness & Relevancy \\ 
        \midrule
        \multirow{2}{*}{Llama-2-7B} & $\times$ & 0.237 & 3.938 & 0.985 & 0.336 & 3.764 & 0.982 \\ 
                                      & $\checkmark$ & \textbf{0.239} & \textbf{3.974} & \textbf{0.987} & \textbf{0.340} & \textbf{3.847} & \textbf{0.984} \\ 
        \midrule
        \multirow{2}{*}{Llama-2-13B}    & $\times$ & 0.262 & 4.282 & 0.990 & 0.324 & 4.148 & 0.989 \\ 
                                      & $\checkmark$ & \textbf{0.263} & \textbf{4.295} & \textbf{0.994} & \textbf{0.325} & \textbf{4.161} & \textbf{0.997} \\
        \bottomrule
    \end{tabular}
    \caption{RAG performance on ELI5 and ASQA, with both automatic evaluation and GPT evaluation.}
    \label{tab:rag}
    
\end{table*}

We further test the performance of entailment tuning on open-domain QA, a downstream task where the answer should be entailed in the retrieved passage as we previously analyzed in Sec.~\ref{sec:3}.

In open-domain QA task, the retriever first retrieves relevant passages from a large corpus given a query. Then, a reader will comprehend the content of retrieved passages and extract or generate the final answer.

We use a widely used strong reader FiD~\cite{fid2021} in the reading comprehension part. It pairs query with each passage, and uses a fused representation of all retrieved passages to decode the final answer. Following FiD, we test our method using both base and large T5~\cite{t52020} models and use exact match(EM) as evaluation metric. Results in Table~\ref{tab:odqa} show that entail tuning improves the accuracy of answers.

\subsection{RAG}
Different from traditional QA, RAG utilizes the generation power of large language models to deal with complex generation tasks, such as long-form question answering, code generation, and task implementation. To test whether the entailment relationship benefits relevant tasks in RAG, we test our method on ELI5~\cite{dataset:eli52019} and ASQA~\cite{dataset:asqa2022}, two long-form answer generation dataset. For the generator, we use LLaMA-2-7B and 13B models to generate responses based on our retrieved passages. 

\noindent\textbf{Automatic Evaluation}. We first measure the quality of response using ROUGE score. ROUGE score calculate a pairwise similarity of the generation to the groundtruth reference, with a higher score indicate better alignment with the groundtruth. 

\begin{figure}[h]
    \centering
    \includegraphics[width=0.5\textwidth]{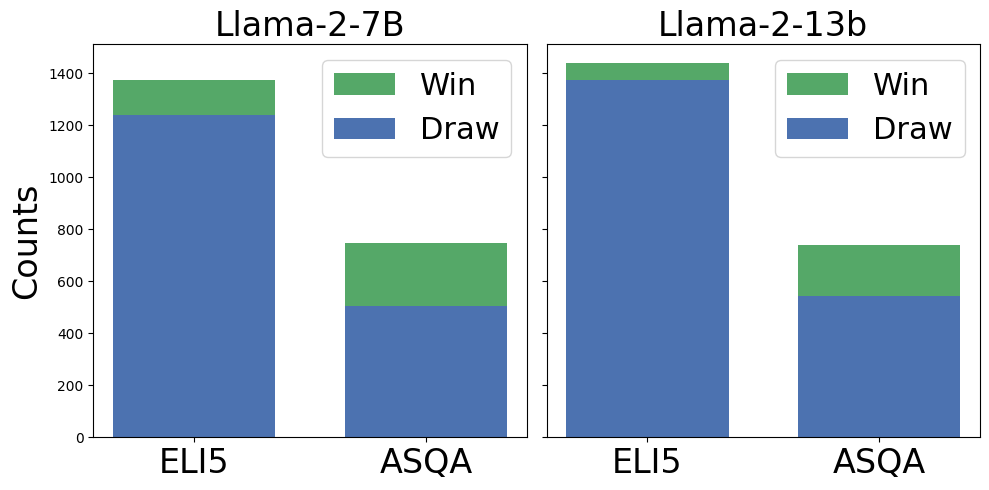}
    \caption{Pairwise Comparison by GPT-4. Our method wins over or tie with baselines in general quality.}
    \label{fig:gpt-pairwise}
\end{figure}

\noindent\textbf{Human-based Evaluation}. While statistic-based metric ROUGE score can assess generation results based on lexical matching, it cannot cover complex aspects such as helpfulness, fluency and correctness, which reflects the true quality of RAG\cite{rougehurdle2021}. To evaluate RAG results from diverse dimensions, we also employ GPT-4 as evaluators to mimic human beings in assessing the quality of generation. Specifically, we follow LlamaIndex\cite{Liu_LlamaIndex_2022} and use the correctness, answer relevancy and pairwise score as quality criterion, elaborated in Appendix. Correctness is a 1-5 score indicating the level of responses' faithfulness to truth. Answer relevancy is a 0/1 score indicating whether response provide helpful answer to the query. Pairwise Score is a 0/1 score given a pair of generations, with 1 indicating the first is better than the second. Results in Table~\ref{tab:rag} and Figure~\ref{fig:gpt-pairwise} shows that our method receives higher scores both in correctness and relevancy compared to baselines on both datasets.

\subsection{Ablations}
We further do ablation experiments on our entailment tuning method. (RQ1) What's the best mask prediction strategy for entailment tuning? (RQ2) Whether unified prompting is a better choice than using a concatenation of passages and questions directly? 
For RQ1, we tested two variants of default a mask ratio $\beta = 0.8$ over hypothesis(H): $\beta= 0.2$ over H, and apply mask over the full prompt(F). 
Results in Table~\ref{tab:ablation} show that applying aggressive mask on hypothesis has a noticable advantage over others.

\begin{table}[h]
  \centering
  \small
  \begin{tabular}{ccccc}
    \toprule
    Exp. Setting & R@1 & R@5 & R@50 & MRR  \\
    \midrule
    -- & 45.21 & 68.20 & 83.82 & 64.51  \\
    \midrule
    \multicolumn{5}{l}{\textit{Mask Strategy}} \\
    \midrule
    $\beta=0.2/H$ & 46.37 & 69.36 & 84.21 & 65.35 \\
    $\beta=0.8/F$ & 46.26 & 68.67 & 84.43 & 65.00\\
    $\beta=0.8/H$ & 48.53 & 70.08 & 84.52 & 67.24  \\
    \midrule
    \multicolumn{5}{l}{\textit{Prompt Strategy}} \\
    \midrule
    $p\ \text{[SEP]}\ q$ & 45.79 & 68.95 & 83.82 & 65.18 \\
    $Prompt(p, c)$ & 48.53 & 70.08 & 84.52 & 67.24  \\    
    \bottomrule
  \end{tabular}
  \caption{Ablation on mask strategy and prompt strategy. $c$ is the existence claim transformed from $q$. [SEP] is the concatenation token in BERT.}
  \label{tab:ablation}
\end{table}

For RQ2, we compare MLM with unified prompt method and MLM with a simple concatenation method. Results show that it is not trivial to transform question into existence claim and use unified natural prompt for MLM training. 
\section{Conclusion}

In this work, we study the definition of relevance in retrieval, especially in the setting of dense retrieval for QA. We bring forward the connection between dense passage retrieval and NLI through an information-invariant question-to-claim transformation trick. Based on this perspective, we conduct logical-form analysis and find experimental evidence to validate its reasonability. We further design an effective and efficient method called entailment tuning which can be easily plugged into the current dense retriever training pipeline. Empirical results on dense passage retrieval and downstream tasks including open-domain QA and RAG prove the advantage of our methods.

\section{Acknowledgements}
We sincerely thank all the reviewers for their valuable suggestions.
This work was supported by the National Key R\&D Program of China (Grant No.2023YFF0725001), National Natural Science Foundation of China (Grant No.92370204, No.62102110), Guangzhou-HKUST(GZ) Joint Funding Program (Grant No.2023A03J0008), CCF-DiDi GAIA Collaborative Research Funds, Education Bureau of Guangzhou Municipality.

\bibliography{main}

\begin{thebibliography}{45}
\providecommand{\natexlab}[1]{#1}

\bibitem[{Asai et~al.(2023)Asai, Schick, Lewis, Chen, Izacard, Riedel, Hajishirzi, and Yih}]{tart2023}
Akari Asai, Timo Schick, Patrick Lewis, Xilun Chen, Gautier Izacard, Sebastian Riedel, Hannaneh Hajishirzi, and Wen-tau Yih. 2023.
\newblock Task-aware retrieval with instructions.
\newblock In \emph{Findings of the Association for Computational Linguistics: ACL 2023}, pages 3650--3675.

\bibitem[{Asai et~al.(2024)Asai, Zhong, Chen, Koh, Zettlemoyer, Hajishirzi, and Yih}]{survey:rag2024}
Akari Asai, Zexuan Zhong, Danqi Chen, Pang~Wei Koh, Luke Zettlemoyer, Hannaneh Hajishirzi, and Wen-tau Yih. 2024.
\newblock Reliable, adaptable, and attributable language models with retrieval.
\newblock \emph{arXiv preprint arXiv:2403.03187}.

\bibitem[{BehnamGhader et~al.(2024)BehnamGhader, Adlakha, Mosbach, Bahdanau, Chapados, and Reddy}]{llm2vec2024}
Parishad BehnamGhader, Vaibhav Adlakha, Marius Mosbach, Dzmitry Bahdanau, Nicolas Chapados, and Siva Reddy. 2024.
\newblock Llm2vec: Large language models are secretly powerful text encoders.
\newblock \emph{arXiv preprint arXiv:2404.05961}.

\bibitem[{Bowman et~al.(2015)Bowman, Angeli, Potts, and Manning}]{dataset:snli2015}
Samuel Bowman, Gabor Angeli, Christopher Potts, and Christopher~D Manning. 2015.
\newblock A large annotated corpus for learning natural language inference.
\newblock In \emph{Proceedings of the 2015 Conference on Empirical Methods in Natural Language Processing}, pages 632--642.

\bibitem[{Bruno and Roth(2022)}]{lawnli2022}
William Bruno and Dan Roth. 2022.
\newblock Lawngnli: A long-premise benchmark for in-domain generalization from short to long contexts and for implication-based retrieval.
\newblock In \emph{Findings of the Association for Computational Linguistics: EMNLP 2022}, pages 5019--5043.

\bibitem[{Cer et~al.(2018)Cer, Yang, Kong, Hua, Limtiaco, John, Constant, Guajardo-Cespedes, Yuan, Tar et~al.}]{use2018}
Daniel Cer, Yinfei Yang, Sheng-yi Kong, Nan Hua, Nicole Limtiaco, Rhomni~St John, Noah Constant, Mario Guajardo-Cespedes, Steve Yuan, Chris Tar, et~al. 2018.
\newblock Universal sentence encoder for english.
\newblock In \emph{Proceedings of the 2018 conference on empirical methods in natural language processing: system demonstrations}, pages 169--174.

\bibitem[{Chen et~al.(2017)Chen, Fisch, Weston, and Bordes}]{drqa2017}
Danqi Chen, Adam Fisch, Jason Weston, and Antoine Bordes. 2017.
\newblock Reading wikipedia to answer open-domain questions.
\newblock In \emph{Proceedings of the 55th Annual Meeting of the Association for Computational Linguistics (Volume 1: Long Papers)}, pages 1870--1879.

\bibitem[{Conneau et~al.(2017)Conneau, Kiela, Schwenk, Barrault, and Bordes}]{stsfromnli2017}
Alexis Conneau, Douwe Kiela, Holger Schwenk, Lo{\"\i}c Barrault, and Antoine Bordes. 2017.
\newblock Supervised learning of universal sentence representations from natural language inference data.
\newblock In \emph{Proceedings of the 2017 Conference on Empirical Methods in Natural Language Processing}, pages 670--680.

\bibitem[{Devlin et~al.(2019)Devlin, Chang, Lee, and Toutanova}]{bert2019}
Jacob Devlin, Ming-Wei Chang, Kenton Lee, and Kristina Toutanova. 2019.
\newblock {BERT}: Pre-training of deep bidirectional transformers for language understanding.
\newblock In \emph{Proceedings of the 2019 Conference of the North {A}merican Chapter of the Association for Computational Linguistics: Human Language Technologies, Volume 1 (Long and Short Papers)}.

\bibitem[{Fan et~al.(2019)Fan, Jernite, Perez, Grangier, Weston, and Auli}]{dataset:eli52019}
Angela Fan, Yacine Jernite, Ethan Perez, David Grangier, Jason Weston, and Michael Auli. 2019.
\newblock Eli5: Long form question answering.
\newblock In \emph{Proceedings of the 57th Annual Meeting of the Association for Computational Linguistics}, pages 3558--3567.

\bibitem[{Gao and Callan(2021)}]{condenser2021}
Luyu Gao and Jamie Callan. 2021.
\newblock Condenser: a pre-training architecture for dense retrieval.
\newblock In \emph{Proceedings of the 2021 Conference on Empirical Methods in Natural Language Processing}, pages 981--993.

\bibitem[{Gao et~al.(2023)Gao, Ma, Lin, and Callan}]{hyde2023}
Luyu Gao, Xueguang Ma, Jimmy Lin, and Jamie Callan. 2023.
\newblock Precise zero-shot dense retrieval without relevance labels.
\newblock In \emph{Proceedings of the 61st Annual Meeting of the Association for Computational Linguistics (Volume 1: Long Papers)}, pages 1762--1777.

\bibitem[{Gao et~al.(2021)Gao, Yao, and Chen}]{simcse2021}
Tianyu Gao, Xingcheng Yao, and Danqi Chen. 2021.
\newblock Simcse: Simple contrastive learning of sentence embeddings.
\newblock In \emph{Proceedings of the 2021 Conference on Empirical Methods in Natural Language Processing}, pages 6894--6910.

\bibitem[{Guu et~al.(2020)Guu, Lee, Tung, Pasupat, and Chang}]{realm2020}
Kelvin Guu, Kenton Lee, Zora Tung, Panupong Pasupat, and Mingwei Chang. 2020.
\newblock Retrieval augmented language model pre-training.
\newblock In \emph{International conference on machine learning}, pages 3929--3938. PMLR.

\bibitem[{He et~al.(2020)He, Liu, Gao, and Chen}]{deberta2020}
Pengcheng He, Xiaodong Liu, Jianfeng Gao, and Weizhu Chen. 2020.
\newblock Deberta: Decoding-enhanced bert with disentangled attention.
\newblock In \emph{International Conference on Learning Representations}.

\bibitem[{Humeau et~al.(2019)Humeau, Shuster, Lachaux, and Weston}]{polyencoder2019}
Samuel Humeau, Kurt Shuster, Marie-Anne Lachaux, and Jason Weston. 2019.
\newblock Poly-encoders: Architectures and pre-training strategies for fast and accurate multi-sentence scoring.
\newblock In \emph{International Conference on Learning Representations}.

\bibitem[{Izacard et~al.(2022)Izacard, Caron, Hosseini, Riedel, Bojanowski, Joulin, and Grave}]{contriever2022}
Gautier Izacard, Mathilde Caron, Lucas Hosseini, Sebastian Riedel, Piotr Bojanowski, Armand Joulin, and Edouard Grave. 2022.
\newblock Unsupervised dense information retrieval with contrastive learning.
\newblock \emph{Transactions on Machine Learning Research}.

\bibitem[{Izacard and Grave(2021)}]{fid2021}
Gautier Izacard and Edouard Grave. 2021.
\newblock Leveraging passage retrieval with generative models for open domain question answering.
\newblock In \emph{Proceedings of the 16th Conference of the European Chapter of the Association for Computational Linguistics: Main Volume}.

\bibitem[{Johnson et~al.(2019)Johnson, Douze, and J{\'e}gou}]{faiss2019}
Jeff Johnson, Matthijs Douze, and Herv{\'e} J{\'e}gou. 2019.
\newblock Billion-scale similarity search with gpus.
\newblock \emph{IEEE Transactions on Big Data}, 7(3):535--547.

\bibitem[{Karpukhin et~al.(2020)Karpukhin, Oguz, Min, Lewis, Wu, Edunov, Chen, and Yih}]{dpr2020}
Vladimir Karpukhin, Barlas Oguz, Sewon Min, Patrick Lewis, Ledell Wu, Sergey Edunov, Danqi Chen, and Wen-tau Yih. 2020.
\newblock Dense passage retrieval for open-domain question answering.
\newblock In \emph{Proceedings of the 2020 Conference on Empirical Methods in Natural Language Processing (EMNLP)}.

\bibitem[{Ke et~al.(2024)Ke, Kong, Li, Zhang, Mei, and Bendersky}]{bridgegap2024}
Zixuan Ke, Weize Kong, Cheng Li, Mingyang Zhang, Qiaozhu Mei, and Michael Bendersky. 2024.
\newblock Bridging the preference gap between retrievers and llms.
\newblock \emph{arXiv preprint arXiv:2401.06954}.

\bibitem[{Khattab and Zaharia(2020)}]{colbert2020}
Omar Khattab and Matei Zaharia. 2020.
\newblock Colbert: Efficient and effective passage search via contextualized late interaction over bert.
\newblock In \emph{Proceedings of the 43rd International ACM SIGIR conference on research and development in Information Retrieval}, pages 39--48.

\bibitem[{Kim(2014)}]{cnnsts2014}
Yoon Kim. 2014.
\newblock Convolutional neural networks for sentence classification.
\newblock In \emph{Proceedings of the 2014 Conference on Empirical Methods in Natural Language Processing ({EMNLP})}.

\bibitem[{Krishna et~al.(2021)Krishna, Roy, and Iyyer}]{rougehurdle2021}
Kalpesh Krishna, Aurko Roy, and Mohit Iyyer. 2021.
\newblock Hurdles to progress in long-form question answering.
\newblock In \emph{Proceedings of the 2021 Conference of the North American Chapter of the Association for Computational Linguistics: Human Language Technologies}, pages 4940--4957.

\bibitem[{Kwiatkowski et~al.(2019)Kwiatkowski, Palomaki, Redfield, Collins, Parikh, Alberti, Epstein, Polosukhin, Devlin, Lee, Toutanova, Jones, Kelcey, Chang, Dai, Uszkoreit, Le, and Petrov}]{dataset:nq2019}
Tom Kwiatkowski, Jennimaria Palomaki, Olivia Redfield, Michael Collins, Ankur Parikh, Chris Alberti, Danielle Epstein, Illia Polosukhin, Jacob Devlin, Kenton Lee, Kristina Toutanova, Llion Jones, Matthew Kelcey, Ming-Wei Chang, Andrew~M. Dai, Jakob Uszkoreit, Quoc Le, and Slav Petrov. 2019.
\newblock Natural questions: A benchmark for question answering research.
\newblock \emph{Transactions of the Association for Computational Linguistics}, 7.

\bibitem[{Lee et~al.(2019)Lee, Chang, and Toutanova}]{ict2019}
Kenton Lee, Ming-Wei Chang, and Kristina Toutanova. 2019.
\newblock Latent retrieval for weakly supervised open domain question answering.
\newblock In \emph{Proceedings of the 57th Annual Meeting of the Association for Computational Linguistics}, pages 6086--6096.

\bibitem[{Lewis et~al.(2020)Lewis, Perez, Piktus, Petroni, Karpukhin, Goyal, K{\"u}ttler, Lewis, Yih, Rockt{\"a}schel et~al.}]{rag2020}
Patrick Lewis, Ethan Perez, Aleksandra Piktus, Fabio Petroni, Vladimir Karpukhin, Naman Goyal, Heinrich K{\"u}ttler, Mike Lewis, Wen-tau Yih, Tim Rockt{\"a}schel, et~al. 2020.
\newblock Retrieval-augmented generation for knowledge-intensive nlp tasks.
\newblock \emph{Advances in Neural Information Processing Systems}, 33:9459--9474.

\bibitem[{Li et~al.(2020)Li, Zhou, He, Wang, Yang, and Li}]{bertflow2020}
Bohan Li, Hao Zhou, Junxian He, Mingxuan Wang, Yiming Yang, and Lei Li. 2020.
\newblock On the sentence embeddings from pre-trained language models.
\newblock In \emph{Proceedings of the 2020 Conference on Empirical Methods in Natural Language Processing (EMNLP)}, pages 9119--9130.

\bibitem[{Liu(2022)}]{Liu_LlamaIndex_2022}
Jerry Liu. 2022.
\newblock \href {https://doi.org/10.5281/zenodo.1234} {{LlamaIndex}}.

\bibitem[{Liu et~al.(2019)Liu, Ott, Goyal, Du, Joshi, Chen, Levy, Lewis, Zettlemoyer, and Stoyanov}]{roberta2019}
Yinhan Liu, Myle Ott, Naman Goyal, Jingfei Du, Mandar Joshi, Danqi Chen, Omer Levy, Mike Lewis, Luke Zettlemoyer, and Veselin Stoyanov. 2019.
\newblock Roberta: A robustly optimized bert pretraining approach.
\newblock \emph{arXiv preprint arXiv:1907.11692}.

\bibitem[{Ni et~al.(2022)Ni, Abrego, Constant, Ma, Hall, Cer, and Yang}]{st52022}
Jianmo Ni, Gustavo~Hernandez Abrego, Noah Constant, Ji~Ma, Keith Hall, Daniel Cer, and Yinfei Yang. 2022.
\newblock Sentence-t5: Scalable sentence encoders from pre-trained text-to-text models.
\newblock In \emph{Findings of the Association for Computational Linguistics: ACL 2022}, pages 1864--1874.

\bibitem[{Ostendorff et~al.(2022)Ostendorff, Rethmeier, Augenstein, Gipp, and Rehm}]{scincl2022}
Malte Ostendorff, Nils Rethmeier, Isabelle Augenstein, Bela Gipp, and Georg Rehm. 2022.
\newblock Neighborhood contrastive learning for scientific document representations with citation embeddings.
\newblock In \emph{Proceedings of the 2022 Conference on Empirical Methods in Natural Language Processing}, pages 11670--11688.

\bibitem[{Qu et~al.(2021)Qu, Ding, Liu, Liu, Ren, Zhao, Dong, Wu, and Wang}]{rocketqa2021}
Yingqi Qu, Yuchen Ding, Jing Liu, Kai Liu, Ruiyang Ren, Wayne~Xin Zhao, Daxiang Dong, Hua Wu, and Haifeng Wang. 2021.
\newblock Rocketqa: An optimized training approach to dense passage retrieval for open-domain question answering.
\newblock In \emph{Proceedings of the 2021 Conference of the North American Chapter of the Association for Computational Linguistics: Human Language Technologies}, pages 5835--5847.

\bibitem[{Raffel et~al.(2020)Raffel, Shazeer, Roberts, Lee, Narang, Matena, Zhou, Li, and Liu}]{t52020}
Colin Raffel, Noam Shazeer, Adam Roberts, Katherine Lee, Sharan Narang, Michael Matena, Yanqi Zhou, Wei Li, and Peter~J. Liu. 2020.
\newblock Exploring the limits of transfer learning with a unified text-to-text transformer.
\newblock \emph{Journal of Machine Learning Research}, 21(140):1--67.

\bibitem[{Reimers and Gurevych(2019)}]{sbert2019}
Nils Reimers and Iryna Gurevych. 2019.
\newblock Sentence-{BERT}: Sentence embeddings using {S}iamese {BERT}-networks.
\newblock In \emph{Proceedings of the 2019 Conference on Empirical Methods in Natural Language Processing and the 9th International Joint Conference on Natural Language Processing (EMNLP-IJCNLP)}.

\bibitem[{Robertson and Zaragoza(2009)}]{bm252009}
Stephen Robertson and Hugo Zaragoza. 2009.
\newblock The probabilistic relevance framework: Bm25 and beyond.
\newblock \emph{Found. Trends Inf. Retr.}

\bibitem[{Stelmakh et~al.(2022)Stelmakh, Luan, Dhingra, and Chang}]{dataset:asqa2022}
Ivan Stelmakh, Yi~Luan, Bhuwan Dhingra, and Ming-Wei Chang. 2022.
\newblock Asqa: Factoid questions meet long-form answers.
\newblock In \emph{Proceedings of the 2022 Conference on Empirical Methods in Natural Language Processing}, pages 8273--8288.

\bibitem[{Su et~al.(2023)Su, Shi, Kasai, Wang, Hu, Ostendorf, Yih, Smith, Zettlemoyer, and Yu}]{instructor2023}
Hongjin Su, Weijia Shi, Jungo Kasai, Yizhong Wang, Yushi Hu, Mari Ostendorf, Wen-tau Yih, Noah~A Smith, Luke Zettlemoyer, and Tao Yu. 2023.
\newblock One embedder, any task: Instruction-finetuned text embeddings.
\newblock In \emph{Findings of the Association for Computational Linguistics: ACL 2023}, pages 1102--1121.

\bibitem[{Thorne et~al.(2018)Thorne, Vlachos, Cocarascu, Christodoulopoulos, and Mittal}]{fever2018}
James Thorne, Andreas Vlachos, Oana Cocarascu, Christos Christodoulopoulos, and Arpit Mittal. 2018.
\newblock The fact extraction and {VER}ification ({FEVER}) shared task.
\newblock In \emph{Proceedings of the First Workshop on Fact Extraction and {VER}ification ({FEVER})}.

\bibitem[{Van~Opijnen and Santos(2017)}]{lawrelevance2017}
Marc Van~Opijnen and Cristiana Santos. 2017.
\newblock On the concept of relevance in legal information retrieval.
\newblock \emph{Artificial Intelligence and Law}, 25:65--87.

\bibitem[{Wang et~al.(2024)Wang, Wang, Wang, Naidu, Bergen, and Paturi}]{dorismae2024}
Jianyou~Andre Wang, Kaicheng Wang, Xiaoyue Wang, Prudhviraj Naidu, Leon Bergen, and Ramamohan Paturi. 2024.
\newblock Scientific document retrieval using multi-level aspect-based queries.
\newblock \emph{Advances in Neural Information Processing Systems}, 36.

\bibitem[{Wettig et~al.(2023)Wettig, Gao, Zhong, and Chen}]{shouldmask2023}
Alexander Wettig, Tianyu Gao, Zexuan Zhong, and Danqi Chen. 2023.
\newblock Should you mask 15\% in masked language modeling?
\newblock In \emph{Proceedings of the 17th Conference of the European Chapter of the Association for Computational Linguistics}, pages 2985--3000.

\bibitem[{Williams et~al.(2018)Williams, Nangia, and Bowman}]{dataset:mnli2018}
Adina Williams, Nikita Nangia, and Samuel Bowman. 2018.
\newblock A broad-coverage challenge corpus for sentence understanding through inference.
\newblock In \emph{Proceedings of the 2018 Conference of the North American Chapter of the Association for Computational Linguistics: Human Language Technologies, Volume 1 (Long Papers)}, pages 1112--1122.

\bibitem[{Xiao et~al.(2022)Xiao, Liu, Shao, and Cao}]{retromae2022}
Shitao Xiao, Zheng Liu, Yingxia Shao, and Zhao Cao. 2022.
\newblock Retromae: Pre-training retrieval-oriented language models via masked auto-encoder.
\newblock In \emph{Proceedings of the 2022 Conference on Empirical Methods in Natural Language Processing}, pages 538--548.

\bibitem[{Xiong et~al.(2020)Xiong, Xiong, Li, Tang, Liu, Bennett, Ahmed, and Overwijk}]{ance2020}
Lee Xiong, Chenyan Xiong, Ye~Li, Kwok-Fung Tang, Jialin Liu, Paul~N Bennett, Junaid Ahmed, and Arnold Overwijk. 2020.
\newblock Approximate nearest neighbor negative contrastive learning for dense text retrieval.
\newblock In \emph{International Conference on Learning Representations}.

\end{thebibliography}

\section{Limitations}
While our work provides some insights on a more concrete definition of relevance, it has several limitations. First, the entailment relationship can only accurately capture the relevance in QA-related retrieval. As shown in Table~\ref{tab:relevance_type}, there exists other types of user intent in retrieval, such as retrieve contradictory opinions and satisfy user instructions. To gain better understanding of relevance and improve the general retrieval performance, it's important to examine and investigate into different types of relevance cases in future research. Second, our method works in a dense retrieval setting. Since NLI requires high-level semantic understanding of texts, it's hard to use sparse retrieval methods which heavily rely on lexical similarity to discerning entailment relationship between passages and claims. This also motivates us to design build our entailment method on modern PLMs.

\appendix

\section{Appendix}
\label{sec:appendix}
\subsection{Experimental Details}

A dense retriever training can be roughly divided into two stages: retrieval-oriented pre-train and constrastive based fine-tune. Our entailment tuning come in between the two stages. In our method, a PLM is first trained using our entailment-tuning method, followed by existing contrastive fine-tuning methods such as DPR and ANCE. 

Dataset used in entailment tuning includes NQ/MSMARCO, SNLI and MNLI. We train 10 epochs on 8 A6000 49G GPUs, which costs around 1.5 hours to finish for NQ setting and 3.5 hours for MSMARCO setting. The statistics of NQ and MSMARCO are listed below in Table~\ref{tab:dataset_stats}. Training parameters for entailment tuning is elaborated in Table~\ref{tab:parameter}.

At the fine-tuning stage, we follow the exact training hyper-parameters as DPR\cite{dpr2020}. The passage a chunk of 100 words and with a limited token number 256. Fine-tuning consists of 40 epochs in NQ and 2 epochs on MSMARCO given the size of MSMARCO corpus is 10 times larger than NQ.

\begin{table}[ht]
\centering
\begin{tabular}{@{}ll@{}}
\toprule
\textbf{Parameter} & \textbf{Value} \\ 
\midrule
learning\_rate & 2e-5 \\
warmup\_steps & 100 \\
batch\_size & 128 \\
train\_epochs & 10 \\
weight\_decay & 0.01 \\
adam\_beta & (0.9, 0.999) \\
adam\_epsilon & 1e-8 \\
max\_grad\_norm & 1.0 \\
\bottomrule
\end{tabular}
\caption{Training arguments for entailment tuning.}
\label{tab:parameter}
\end{table}

\begin{table}[h]
\small
\centering
\begin{tabular}{@{}lllll@{}}
\toprule
{Dataset} & { Train } & { Dev } & { Test } & { Passage } \\
\midrule
MSMARCO  & 502,939 & 6,980 & 6,837 & 8,841,823 \\
NQ & 58,812  & 6,515 & 3,610 & 21,015,324 \\
\bottomrule
\end{tabular}
\caption{Statistics of NQ and MSMARCO.}
\label{tab:dataset_stats}
\end{table}

\subsection{Inference}

We use FAISS\cite{faiss2019} to build the index for retrieval. Wikipedia corpus costs 65G memory and MSMARCO costs 27G memory. We shard the vector store of corpus into 8 GPUs and use FAISS to organize them. It costs less than 1ms to retrieve top-100 passages for each query.

\subsection{Prompt used for RAG evaluation}.

We follow the default evaluation pipeline in LlamaIndex to evaluate the result of our RAG systems. In particular, we assess the quality of responses using the CorrectnessEvaluator and AnswerRelevancyEvaluator from different aspects. We also use PairwiseComparisonEvaluator to compare the overall quality of responses from retriever with and without entailment tuning. The default prompt templates are listed in Table~\ref{tab:llama_index_template}.

\begin{table*}[ht]
\centering
\captionsetup{skip=10pt}  
\begin{tabular}{p{4cm} p{10cm}}  
\toprule
\textbf{Evaluation Metric} & \textbf{Prompt Template}\\
\midrule
Correctness &  You are an expert evaluation system for a question answering chatbot.

You are given the following information:
- a user query, and
- a generated answer

You may also be given a reference answer to use for reference in your evaluation.

Your job is to judge the relevance and correctness of the generated answer.
Output a single score that represents a holistic evaluation.
You must return your response in a line with only the score.
Do not return answers in any other format.
On a separate line provide your reasoning for the score as well.

Follow these guidelines for scoring:
- Your score has to be between 1 and 5, where 1 is the worst and 5 is the best.
- If the generated answer is not relevant to the user query,
you should give a score of 1.
- If the generated answer is relevant but contains mistakes,
you should give a score between 2 and 3.
- If the generated answer is relevant and fully correct,
you should give a score between 4 and 5.

Example Response:
4.0
The generated answer has the exact same metrics as the reference answer,
    but it is not as concise. \\  
\midrule
Answer Relevancy &  Your task is to evaluate if the response is relevant to the query.
The evaluation should be performed in a step-by-step manner by answering the following questions:
1. Does the provided response match the subject matter of the user's query?
2. Does the provided response attempt to address the focus or perspective on the subject matter taken on by the user's query?
Each question above is worth 1 point. Provide detailed feedback on response according to the criteria questions above  After your feedback provide a final result by strictly following this format: '[RESULT] followed by the integer number representing the total score assigned to the response'

Query: 
 \{query\}
Response: 
 \{response\}
Feedback:\\
\midrule
Pairwise Comparison & Please act as an impartial judge and evaluate the quality of the responses provided by two AI question-answering assistants to the user question perhaps with added reference which are displayed below. You should choose the assistant that follows the user’s instructions and answers the user’s question better using the provided context. Your evaluation should consider factors such as the helpfulness, relevance, accuracy, depth, creativity, and level of detail of their responses. Begin your evaluation by comparing the two responses and provide a short explanation. Avoid any position biases and ensure that the order in which the responses were presented does not influence your decision. Do not allow the length of the responses to influence your evaluation. Do not favor certain names of the assistants. Be as objective as possible. After providing your explanation, output your final verdict by strictly following this format: '[[A]]' if assistant A is better, '[[B]]' if assistant B is better, and '[[C]]' for a tie.
 \\
\bottomrule
\end{tabular}
\caption{Default prompt template in LlamaIndex used in our RAG evaluation setting.}
\label{tab:llama_index_template}
\end{table*}

\end{document}